\pdfoutput=1

\documentclass[11pt]{article}

\usepackage[]{ACL2023}

\usepackage{times}
\usepackage{latexsym}
\usepackage{graphicx}

\usepackage[T1]{fontenc}

\usepackage[utf8]{inputenc}

\usepackage{microtype}

\usepackage{inconsolata}

\title{Neural Machine Translation Data Generation and Augmentation using ChatGPT}

\author{ \bf{Wayne Yang} \\
  University of British Columbia \\
  \texttt{wyang001@student.ubc.ca} \\\And
  Garrett Nicolai \\
  University of British Columbia \\
  \texttt{garrett.nicolai@ubc.ca} \\}

\begin{document}
\maketitle
\begin{abstract}

Neural models have revolutionized the field of machine translation, but creating parallel corpora is expensive and time-consuming.  We investigate an alternative to manual parallel corpora - hallucinated parallel corpora created by generative language models.  Although these models are themselves trained on parallel data, they can leverage a multilingual vector space to create data, and may be able to supplement small manually-procured corpora.  Our experiments highlight two key findings - despite a lack of diversity in their output, the hallucinated data improves the translation signal, even when the domain clashes with the original dataset.
\end{abstract}

\section{Introduction}

\begin{figure}
  \centering
    \includegraphics[width=0.45\textwidth]{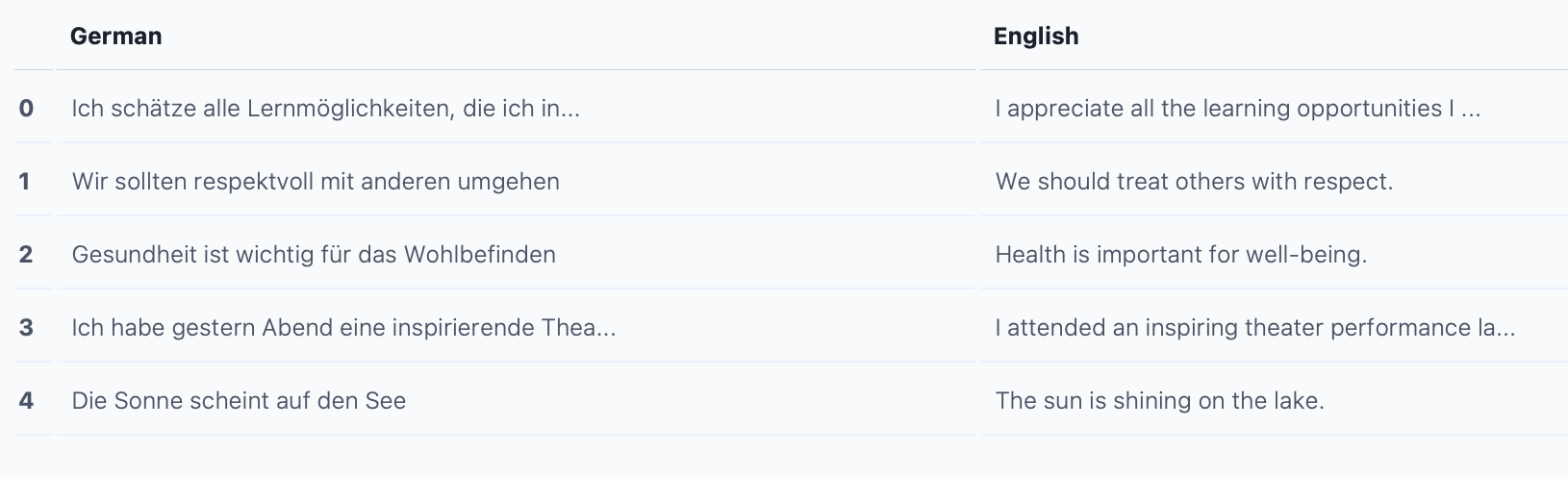}
  \caption{An example of a five parallel German-English sentences.}
  \label{fig:parallel}
\end{figure}

Neural Machine Translation (NMT) models are usually trained on large amounts of sentence-aligned parallel corpora (Figure \ref{fig:parallel}). However, the time and manpower required to create such corpora are quite costly. The same can be said for other Natural Language Processing (NLP) tasks as well, which can explain the reason we have seen increased interest in research on textual data augmentation \citep{feng-etal-2021-survey}. Data augmentation seeks to strengthen the learning objective by providing additional training instances, often generated artificially.

Our augmentation method, commonly referred to as ``data hallucination'' generates a sizable amount of synthetic data independently of the original data \cite{raunak-etal-2021-curious}.  This data is then appended to the original training data in an effort to create a more generalizable dataset. 

In this paper, we propose and explore a simple method for translation data hallucination using ChatGPT \cite{brown2020language}, a prompt-based large language model which has been fine tuned for conversational dialogue. InstructGPT, the precursor to  ChatGPT, has previously been evaluated on Machine Translation \cite{ouyang2022training}, but the full capabilities of ChatGPT in the realm of machine translation are unknown. In this study, we investigate one possible use of ChatGPT for NMT - providing extra data for lower-resourced language pairs.

This paper is structured as follows: In Section \ref{sec:RW}, we cover recent work on low-resource NMT, as well as a preliminary study on ChatGPT; in Section \ref{sec:collection}, we explain our data collection pipeline; Section \ref{sec:experiments} explains how we train our models; our results and interpretation are presented in Section \ref{sec:results}; we provide a qualitative discussion of the data in Section \ref{sec:analysis}, while Section \ref{sec:limitations} addresses the limitations and challenges of this study. Section \ref{sec:conclusions}, concludes the paper. 

\section{Related Work}
\label{sec:RW}

Data augmentation for low-resource NMT often attempts to supplement bilingual data with data from related languages, in an attempt to exploit a cross-lingual signal.  \newcite{mueller-etal-2020-analysis} found that the addition of additional languages can aid low-resource translation, but that the model quickly collapses with too many languages. \newcite{ko-etal-2021-adapting} augmented a low-resource NMT model with monolingual data using their method NMT-Adapt and found improvements over other augmentation methods. \newcite{xia-etal-2019-generalized} make use of a related high-resource language as a pivot to generate synthetic low-resource data \citep{xia-etal-2019-generalized}, but observes that pivoting doesn't perform as well as just augmenting with data of the related high-resource language in most cases. A limitation of these methods is the heavy reliance on the relationships between languages - many low-resource languages do not have higher-resource sisters, and even when they do, cross-lingual transfer differs significantly, even within families. This makes it more difficult to generalize their consistencies across other low-resource languages.

Alternatives to multilingual data augmentation come in several flavors.
Back-translation trains a model that reverses the source and target in the training data. It then feeds monolingual target-side (now- source-side) data to generate parallel sentence. These sentences are then concatenated to the original corpus \citep{sennrich-etal-2016-improving}. \newcite{peng2020dictionarybased} instead proposed a dictionary based approach in addition to back-translation, which showed improvements on cross domain translation. \newcite{ng-etal-2020-ssmba} propose Self-Supervised Manifold Based Data Augmentation (SSMBA) for improving out-of-domain robustness of textual data. In SSMBA, data is sampled from a denoising autoencoder to produce slight variations to the original data that allows for more robust generalization. Other methods, such as word DropOut \citep{sennrich-etal-2016-edinburgh} and SwitchOut \citep{wang-etal-2018-switchout}, apply regularization techniques to limit overfitting. 

ChatGPT is a relatively new toy for MT researchers to play with.
A preliminary study by \newcite{jiao2023chatgpt} discovered that GPT-3 (the underlying language model for ChatGPT) produced fluent-enough results, but substituting GPT-4 produced output comparable to existing commercial NMT systems. Unfortunately, their test size was restricted to 50 sentences, due to a necessity of entering each sentence as a manual prompt. However, the reported quality of the translation inspires our own investigation into the use of ChatGPT for data hallucination.

\section{Data Collection}
\label{sec:collection}

We limit our investigation to two languages: German (de), and Galician (gl).  These languages were chosen due to the amount of data used in the training of GPT-3, ChatGPT's core model.  German can be considered a high-resource language - GPT-3 has seen more than 2.9 billion tokens of German data.  Meanwhile, Galician is a much more sparsely-resourced language, with only 7 million training tokens of Galician data \citep{brown2020language}\footnote{We acknowledge that true low-resource settings will often have orders of magnitude less data than either of these language pairs.}

We use ChatGPT with the GPT-3 language model, accessed through the API (See Section \ref{subsec:api}).  Although a GPT-4 model was available, its API was not yet accessible.
All experiments translate from the source language into English.
\newcite{brown2020language} demonstrate that the GPT model generally performs better with English as a target than as a source.  Furthermore, it allows for a more thorough error analysis and discussion, as none of the authors speak Galician.  

We describe two data settings: Natural data (nat) consists of a corpus extracted from the TEDTalks set \citet{reimers-gurevych-2020-making}.  We randomly select sentences until we surpass a threshold of 1,000,000 tokens in the source language: 900,000 is given to the training set and 100,000 to the validation set.  We likewise sample an additional 100,000 tokens as a test set in each language.

\subsection{Synthetic Data Collection}
\label{subsec:synthetic}

\begin{figure}
  \centering
    \includegraphics[width=0.45\textwidth]{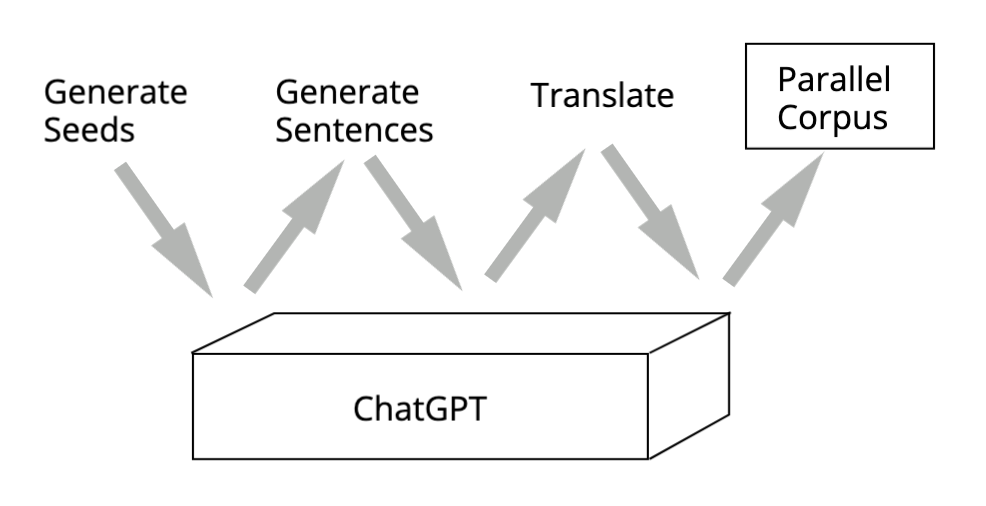}
  \caption{Our process of generating parallel corpora with ChatGPT.  Translated sentences are generated from seed sentences, which themselves are generated from randomly produced seed words.}
  \label{fig:generation}
\end{figure}

In contrast to sampling the natural dataset from an existing corpus, we generate our synthetic (syn) dataset using a three-step process, outlined in Figure \ref{fig:generation}. We first prompt ChatGPT to generate 600 nouns and 600 verbs in our target language; duplicates are removed.  We next prompt ChatGPT to create 100 sentences for each of these seeds. Likewise, duplicates are discarded. Finally, we ask ChatGPT to translate each of our generated sentences into English.  From this corpus, we select enough sentences to have 900,000 training tokens, and 100,000 validation tokens, in a manner that mirrors the sampling of the natural dataset.

\subsubsection{ChatGPT API}
\label{subsec:api}

We prompt ChatGPT using the Python3 API\footnote{https://platform.openai.com/docs/guides/chat}, which also automatically parses and organizes the responses. Prompts for seed words and source sentences are given in the source language, translated using Google Translate\footnote{https://translate.google.com/}, while the translation prompts are given in English. Some example prompts from German can be seen in Table \ref{tab:generation}.

\begin{table}
\centering
\begin{tabular}{p{0.1\textwidth}p{0.33\textwidth}}
\hline
\textbf{Role} & \textbf{Seed Generation Text}\\
\hline
{System} & {`Generieren Sie 600 einzigartige zufällige Substantive, die jeweils durch ein Komma getrennt sind'}\\
\end{tabular}

\begin{tabular}{p{0.1\textwidth}p{0.33\textwidth}}
\hline
\textbf{Role} & \textbf{Sentence Generation Text}\\
\hline
{System} & {`Generieren Sie an der Eingabeaufforderung 100 separate Sätze, die durch ein Semikolon getrennt sind'}\\
{User} & {'Eule'}\\ 
{Assistant} & {`Gärten und Terrassen;Tacos sind gut.;'}\\
\end{tabular}

\begin{tabular}{p{0.1\textwidth}p{0.33\textwidth}}
\hline
\textbf{Role} & \textbf{Translation Text}\\
\hline
{System} & {`Translate from German to English'}\\
{User} & {`Eine Eule ruft durch die Nacht.'}\\
\end{tabular}

\caption{German examples of requests sent to the API.  }
\label{tab:generation}
\end{table}

The API allows 3 types of message inputs: `System', `User', and `Assistant'. The System message provides the model with any context it needs to write its response, such as the task it needs to do and the type of role it should take. ChatGPT was originally designed as a chatbot, so its User role is for user messages and its Assistant role for the model's messages. Multiple User and Assistant messages can be sent in one request, and this differentiation allows prior conversations to be sent back to the model for context on the conversation. Consecutive human-written Assistant messages have been used by the developer community as a method of few shot learning, since the API does not require User and Assistant messages to alternate. The model then responds with an additional Assistant message to continue the conversation. 

For the seed generation task we put the entire prompt, such as ``Generate 600 random unique nouns that are separated by a comma'' as the System message.  For the  sentence generation we then provide the prompt ``Generate 100 sentences generated from the prompt, separated by a semicolon.'' as the System message, the seed word as the User message and an Assistant message consisting of a two sentence example. The two sentence Assistant message is used as a few shot example for delimiter formatting, which ensures a smoother response parsing process. An API call was made for each seed word available to maximize the total number of sentences generated. Finally the translation task gives the prompt, in English, as the System message, and the sentence to translate as a User message (see Table \ref{tab:generation}). Each sentence was given its own API call to reduce the likelihood of mismatching the sentences during parsing. This last step took about 1 second per sentence translation, varying depending on server traffic and internet speeds. 

\begin{figure}
  \centering
    \includegraphics[width=0.45\textwidth]{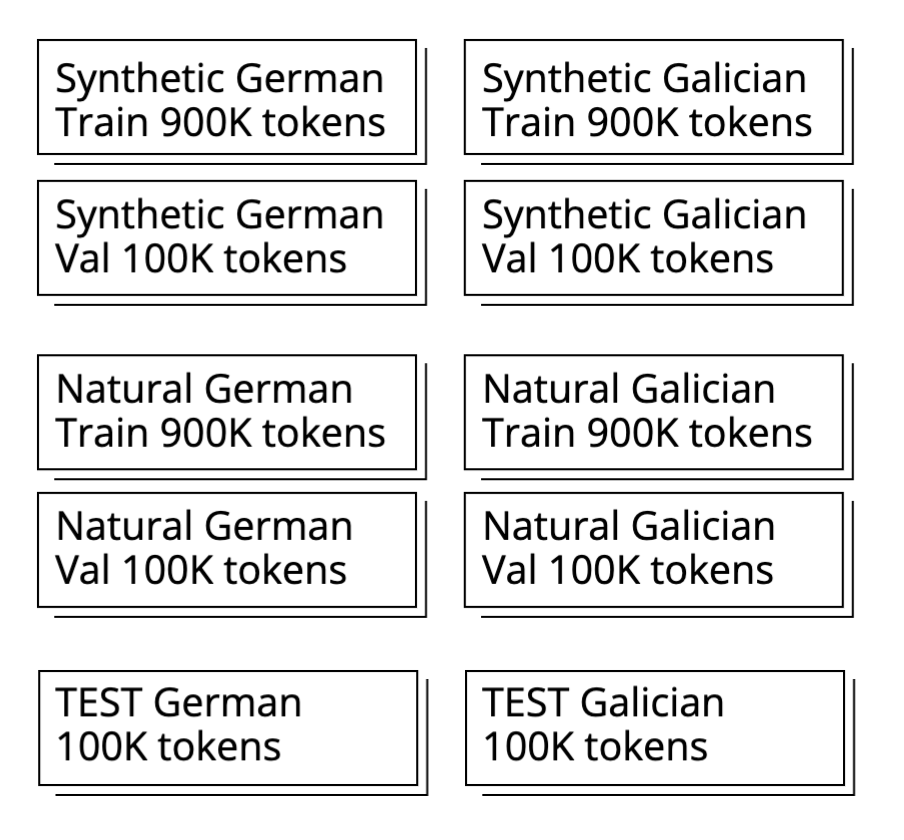}
  \caption{The resulting parallel corpora after our Data Collection section. }
  \label{fig:sets}
\end{figure}

Using this process, we create 10 datasets, outlined in Figure \ref{fig:sets}.  We have a natural training and validation set for both German and Galician.  Likewise, we have a synthetic training and validation set for both languages.  We finally have a natural test set for both languages.

\section{Experimental Design}
\label{sec:experiments}

In our experiment, we consider the role
that synthetically-generated training data can have
on low-resource translation models. We train three
models on each of our language pairs - one that
is trained solely on natural data (Nat), one that is
trained solely on synthetic data (Synth), and one
that is a combination of both (Aug).

Our models are trained using transformers \cite{vaswani2017attention}, implemented using Fairseq \cite{bott-etal-2021-just}.  We train a joint source-target BPE \cite{sennrich-etal-2016-neural} with a vocabulary of size 16K on each language pair.

Models are trained using 4 attention heads and 3 layers, and in a batch size of 2,000.  They are trained for 100 epochs, using validation loss as an early-stopping criterion.  Models are evaluated using SacreBLEU.

\section{Results and Discussion}
\label{sec:results}

To test our six models, we had each of them translate the test set that was encoded using their respective BPE tokenizers. We then ran SacreBLEU on their outputs against the original English sentences from the TED dataset to obtain a BLEU score. The resulting BLEU scores are shown on Table \ref{tab:results}.

We observe that the models trained on natural data are significantly better than those trained on synthetic data.  This is not surprising, as the test set is sampled from the same domain as the natural data.  We did not specify in our prompts to ChatGPT that the generated sentences should be in the same domain as our natural data.

Furthermore, we see that augmenting the natural training data with synthetic data leads to a notable improvement in the translation quality, despite the domain mismatch stated above. In low-resource translation tasks, it appears that any additional information can strengthen the signal.  Even in our "high-resource" setting, the synthetic data supplements the translation.

In Table \ref{tab:qual}, we provide some examples of the predictions made by the augmented model.  We can observe that when the natural and synthetic models are at odds in their predictions (such as in the first example), the augmented model leans towards a correct interpretation.  Likewise, it's able to synergize much more appropriate translations from a combination of the training data.  There is still plenty of room for interpretation and experimentation in how the augmentation is improving translation, which we leave to future work.

\begin{table}
\centering
\begin{tabular}{lc}
\hline
\textbf{Model} & \textbf{Translation}\\
\hline
\textbf{Natural} & {But I am terrible.
} \\
\textbf{Synthetic} & {But I am optimistic.} \\
\textbf{Augmented} & {But I am optimistic.} \\
\textbf{Correct} & \textbf{But I believe in them.
} \\
\hline
\textbf{Natural} & {Vome of spectacular spectrum.} \\
\textbf{Synthetic} & {Defell fell aspect} \\
\textbf{Augmented} & {A lot to speculative.} \\
\textbf{Correct} & \textbf{Highly speculative.} \\
\end{tabular}
\caption{Examples from the German test set. The "Correct" translation is the translation given by the gold standard dataset.}
\label{tab:qual}
\end{table}

\begin{table}
\centering
\begin{tabular}{ccc}
\hline
\textbf{Synth-de} & \textbf{Nat-de} & \textbf{Aug-de}\\
\hline
{3.5} & {16.4} & {18.9}\\
\end{tabular}

\begin{tabular}{ccc}
\hline
\textbf{Synth-gl} & \textbf{Nat-gl} & \textbf{Aug-gl}\\
\hline
{9.5} & {24.4} & {28.3}\\
\end{tabular}
\caption{BLEU scores of the our models on the test set}
\label{tab:results}
\end{table}

Somewhat surprisingly, we observe that although ChatGPT had access to significantly more German data than Galician data, the Galician synthetic data produces higher quality translations than its German counterpart.  Part of this result could just be that Galician is easier to translate into English than German; we observe that in the natural setting, the Galician model produces higher quality translations.  Furthermore, it could be largely coincidental: ChatGPT's training data may better align the limited Galician data \footnote{Indeed, it is possible that GPT-3 was trained on the Galician TEDTalks}. It is also possible that ChatGPT is leveraging a high amount of Spanish data to supplement its Galician translations.  In the next part of our evaluation, we investigate the quality and diversity of the synthetic data.

\begin{table}
\centering
\begin{tabular}{lccc}
\hline
{} & \textbf{Synth-val} & \textbf{Nat-val} & \textbf{Test}\\
\hline
\textbf{Synth-de} & {72.2} & {3.7} & {3.5}\\
\textbf{Synth-gl} & {58.2} & {9.5} & {9.5}\\
\textbf{Nat-de} & {19.0} & {17.0} & {16.4}\\
\textbf{Nat-gl} & {20.3} & {24.4} & {24.4}\\
\textbf{Aug-de} & {-} & {-} & {18.9}\\
\textbf{Aug-gl} & {-} & {-} & {28.3}\\
\end{tabular}
\caption{Cross-method validation BLEU scores.}
\label{diversity}
\label{tab:bleus}
\end{table}

We evaluate our models on both the natural and synthetic validation sets.  As a sanity check, the models perform fairly similarly on natural validation data and natural test data.  However, we note an interesting difference in how the models perform on synthetic data.  The natural models do not provide much insight - the Galician model performs worse on synthetic data, but the German model performs better; further investigation is required.  We note, however, the unnaturally high scores of the synthetic models on the synthetic validation data.  Such high BLEU scores suggest that the synthetic models are overfitting to their validation data - much more than the natural models fit to theirs.  One interpretation of this result could be that the synthetic data is much less diverse than the natural data. Although the training sets are the same size, the model more tightly fits the synthetic data, by a significant margin.

In retrospect, this is not entirely surprising.  ChatGPT is based on a sampled generative language model.  Although it is not as predictable as past language models, it still has a bias towards frequent patterns, and the sentences it is generating may be much more repetitive than we would like. We now turn our investigation to the quality of the data generated by ChatGPT.

\begin{figure*}
  \centering
    \includegraphics[width=0.9\textwidth]{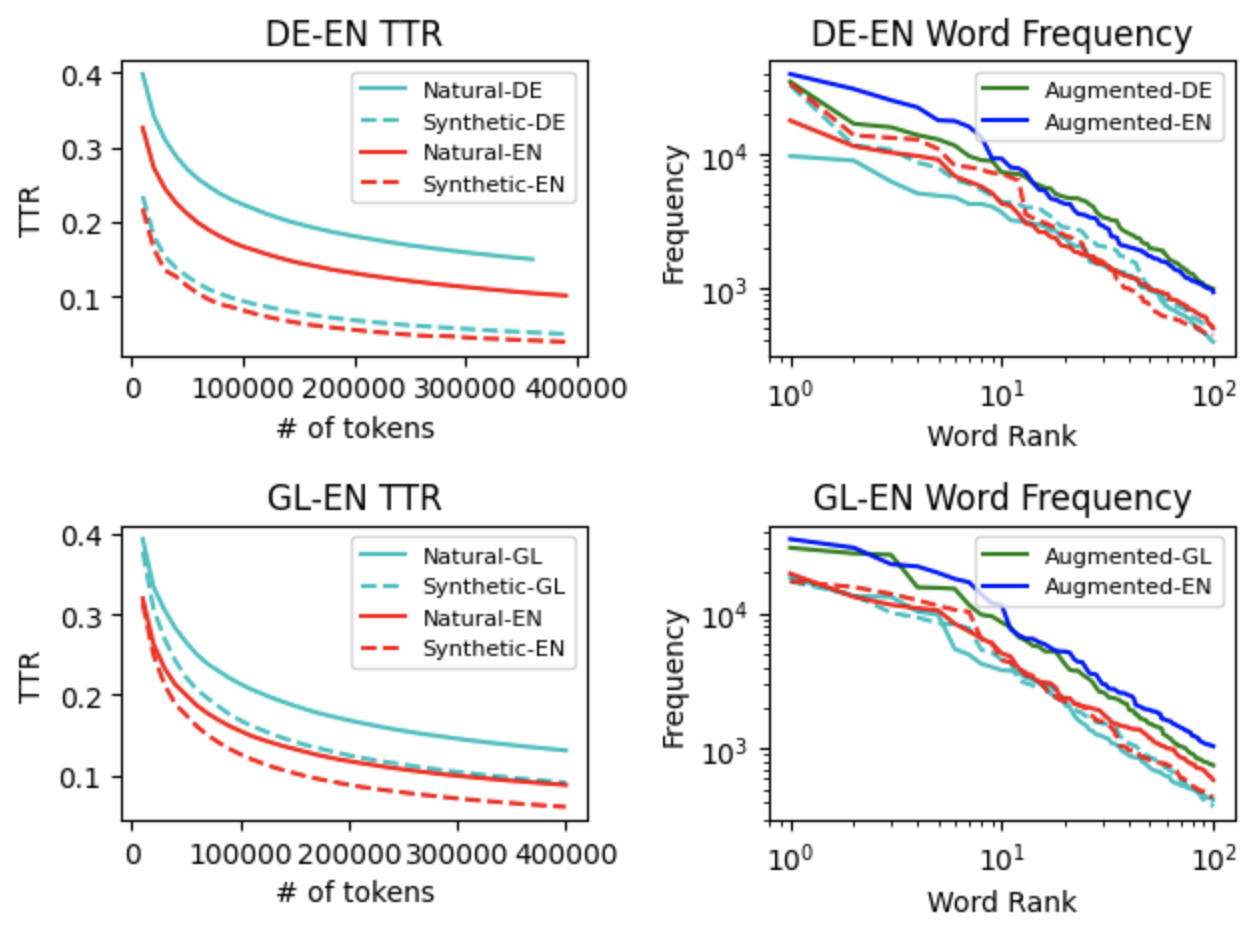}
  \caption{Type token ratio and word frequency distribution plots for the data}
  \label{fig:TTR}
\end{figure*}

\section{Data Analysis}
\label{sec:analysis}

We first investigate the type-token ratio (TTR) of both the natural and synthetic data.  The results are shown on the left of Figure \ref{fig:TTR}.  We see that for both the German and Galician sources, as well as the English targets, the TTR for the synthetic data is much lower than it is for the natural data.  This result confirms our hypothesis that ChatGPT is producing repetitive sentences with a limited vocabulary.  At this point, it is impossible to determine whether this result is a shortcoming of the model's training method, or simply a lack of prompt engineering.

Conversely, we observe that many of the most frequent words in the training data set are used considerably more by the synthetic dataset than the natural one.  Although the synthetic data still follows a roughly Zipfian distribution, its frequent words are regularly more frequent than the natural dataset.  The sentences generated by ChatGPT show significantly less linguistic diversity than the natural sentences.  With a less diverse training set, overfitting is likely.  Since the training and validation sets were sampled from the same distribution, it's highly likely that the synthetic models are also very closely fitting their validation data, as observed in Section \ref{sec:results}.

\section{Limitations}
\label{sec:limitations}

This study is not without limitations. First, the size of the training data we set aside for this study is relatively small for NMT. This may impact the generalizability of our findings to instances where a much larger parallel corpus needs to be augmented. Secondly, our experiments only considered two different languages translating into English (and both languages are western Indo-European languages closely related to English). This is a small sample size and may also affect the generalizability of our study. It is especially important because ChatGPT itself is trained on significantly more data, likely quite a bit of parallel data as well. It's possible it will not perform well in naturally low resource settings, where it is not possible for ChatGPT to generate synthetic parallel data. 

\section{Conclusions}
\label{sec:conclusions}

Our experiments demonstrate a possible issue with using large language models to generate synthetic data: the lack of diversity in generated data.  It is becoming evident that many of the capabilities of these models are limited only by the specific prompts used to communicate with them.  More research is necessary to determine whether prompt engineering can improve the diversity issue. For example, using more sentences during the 'Assistant' few shot phase in section \ref{subsec:api}. As new and varied prompt-based models are released, the problem may also naturally be solved by better language models and more thorough linguistic understanding on the part of the models.

Our experiments do show, however, that augmenting natural data with entirely synthetic data shows promise for training machine translators.  Despite a domain mismatch, translation quality improved, even when the language model only had access to limited training data.  While the diversity issue remains, it is encouraging that even simple, repetitive sentences can improve the quality of a translator.

\section*{Acknowledgements}

We thank Dr. Miikka Silfverberg for providing suggestions in the data analysis section, as well as Dr. James Kryklywy for insight in the discussion of the results.

\nocite{}

\bibliography{anthology,custom}
\bibliographystyle{acl_natbib}

\end{document}